\pdfoutput=1

\documentclass[11pt]{article}

\usepackage[svgnames,table]{xcolor}

\usepackage[final]{acl}

\usepackage{times}
\usepackage{latexsym}
\usepackage{fancyhdr}  

\usepackage[T1]{fontenc}

\usepackage[utf8]{inputenc}

\usepackage{microtype}

\usepackage{inconsolata}

\usepackage{graphicx}
\usepackage{verbatim}
\usepackage{booktabs}  
\usepackage{multicol}
\usepackage{multirow}
\usepackage{tabularx}

\definecolor{mygray}{HTML}{F0F0F0}
\definecolor{myblue}{HTML}{DDEBF3}

\usepackage{CJK}
\usepackage{bbding}
\usepackage{pifont}
\usepackage{fontawesome5}
\usepackage[many]{tcolorbox}
\usepackage{hyperref}

\fancypagestyle{firstpage}{
  \fancyhf{}  
  \lhead{\raisebox{-0.2\height}[0pt][0pt]{%
    \makebox[0pt][l]{%
      \includegraphics[width=0.3\textwidth]{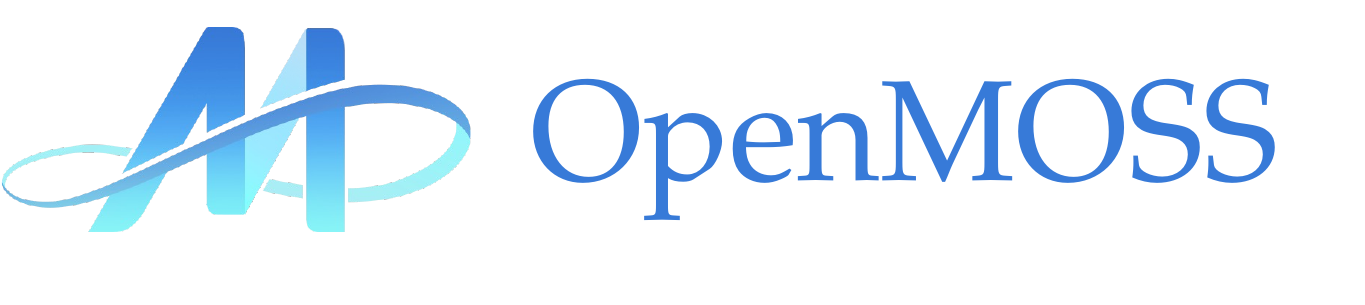}%
    }%
  }}
  
  \setlength{\headwidth}{\textwidth}  
}

\title{\textsc{InstructTTSEval}: Benchmarking Complex Natural-Language Instruction Following in Text-to-Speech Systems}

\author{
Kexin Huang$^{1}$ \hspace{.3em}
Qian Tu$^{1}$ \hspace{.3em}
Liwei Fan$^{1}$ \hspace{.3em}
Chenchen Yang$^{1}$\hspace{.3em}
Dong Zhang$^{1} $
\\
\textbf{
Shimin Li$^{1}$ \hspace{.3em}
Zhaoye Fei$^{1}$ \hspace{.3em}
Qinyuan Cheng$^{1}$ \hspace{.3em}
Xipeng Qiu$^{1,}$\thanks{ {} Corresponding author.}
}
\\
\texttt{\{kxhuang24, chengqy21\}@m.fudan.edu.cn} \quad \texttt{xpqiu@fudan.edu.cn} \\ 
[1ex]
$^{1}$Fudan University \\
}

\begin{document}

\thispagestyle{firstpage}

\maketitle

\pagestyle{plain}

\begin{abstract}

In modern speech synthesis, paralinguistic information—such as a speaker’s vocal timbre, emotional state, and dynamic prosody—plays a critical role in conveying nuance beyond mere semantics. 
Traditional Text-to-Speech~(TTS) systems rely on fixed style labels or inserting a speech prompt to control these cues, which severely limits flexibility. Recent attempts seek to employ natural-language instructions to modulate paralinguistic features, substantially improving the generalization of instruction-driven TTS models. 
Although many TTS systems now support customized synthesis via textual description, their actual ability to interpret and execute complex instructions remains largely unexplored. In addition, there is still a shortage of high-quality benchmarks and automated evaluation metrics specifically designed for instruction-based TTS, which hinders accurate assessment and iterative optimization of these models.
To address these limitations, we introduce \textbf{\textsc{InstructTTSEval}}, a benchmark for measuring the capability of complex natural-language style control. We introduce three tasks, namely Acoustic-Parameter Specification, Descriptive-Style Directive, and Role-Play, including English and Chinese subsets, each with 1k test cases~(6k in total) paired with reference audio. 
We leverage Gemini as an automatic judge to assess their instruction-following abilities. 
Our evaluation of accessible instruction-following TTS systems highlights substantial room for further improvement.
We anticipate that \textsc{InstructTTSEval} will drive progress toward more powerful, flexible, and accurate instruction-following TTS.\footnote{ {} Available at \href{https://github.com/KexinHUANG19/InstructTTSEval}{\faGithub\ InstructTTSEval}. }

\end{abstract}

\section{Introduction}


In recent years, a number of standout TTS systems have emerged~\cite{Ye2025LlasaST, Wang2024MaskGCTZT, fish-speech-v1.4, Ren2020FastSpeech2F, Chen2024F5TTSAF, Betker2023BetterSS, Wang2023NeuralCL, wang2025sparktts, Anastassiou2024SeedTTSAF}, achieving extremely low word- or character-error rates, highly natural “human-like” fluency, and remarkable voice-cloning abilities.

\begin{figure}[tbp]
    \centering
    \includegraphics[width=\linewidth]{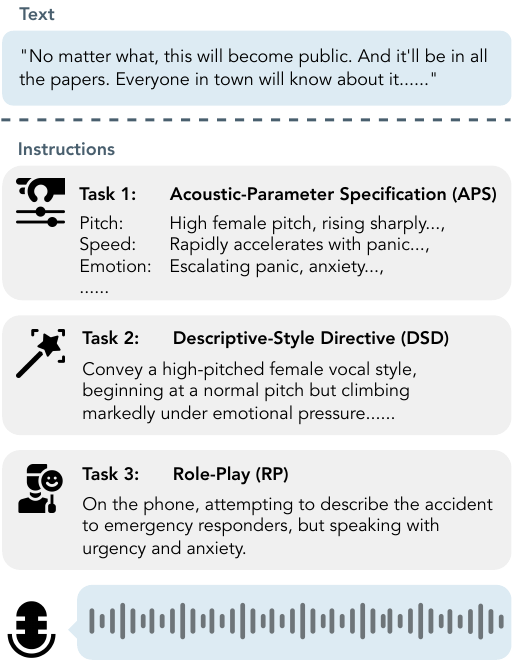}
    \caption{Tasks in \textsc{InstructTTSEval}, progressing from concrete control to abstract expressiveness. APS task evaluates models' accurate control for all low-level acoustic features, DSD task tests a model’s ability to generalize from unstructured prompts, and RP task requires models to infer appropriate vocal styles from high-level character or scenario descriptions.}
    \label{fig:tasks}
\end{figure}

Despite their strong semantic clarity, human conversation carries vital acoustic cues as well~\cite{cutler1997prosody}. In an extreme example, the same text spoken by different people—or by the same person in different moods—can convey entirely different meanings. For instance, when someone says “Help me”, a playful child’s request might simply mean “help me reach that toy” or “tie my shoe,” delivered with innocent enthusiasm rather than real distress; an elderly speaker, however, could express genuine urgency—“I need assistance because I’m injured or unsteady”—signaling true vulnerability rather than casual plea. 
Controlling such acoustic features is crucial even for modern TTS systems: we would not expect comforting words rendered in a cold, detached tone by a speech engine, as that mismatch could largely diminish the quality of the user's experience.



There have been initial attempts to control acoustic features—most use special tokens~(e.g., \texttt{<happy>}) or short phrase prompts~\cite{Du2024CosyVoiceAS, Guo2024FireRedTTSAF, Kim2021ExpressiveTU}, and some recent works explore free-form natural‐language control and show encouraging progress~\cite{Yang2023InstructTTSME, Guo2022PromptttsCT, Leng2023PromptTTS2D, Liu2023PromptStyleCS, Du2024CosyVoice2S, Ji2023TextrolSpeechAT, Ji2024ControlSpeechTS, Lyth2024NaturalLG, Zhou2024VoxInstructEH, Yang2025EmoVoiceLE}. 
However, current metrics are insufficient for evaluation: most common objective metrics measure speech quality like word error rate~(WER) and speaker similarity~(SIM), while subjective MOS evaluations depend on costly human annotation and often suffer from inconsistent standards.
We still lack standardized benchmarks specifically designed to quantify the effectiveness of natural-language instruction-based acoustic control, hindering accurate assessment and iterative model improvement.

To address this gap, we introduce \textbf{\textsc{InstructTTSEval}}, a fully automatic benchmark for measuring a TTS system’s ability to control acoustic features. As shown in Fig.~\ref{fig:tasks}, it consists of three tasks:
1) Acoustic-parameter Specification: models receive a structured set of fine-grained natural-language instructions specifying detailed acoustic characteristics~(e.g., pitch, speed, emotion), and must directly map each descriptive cue to the corresponding acoustic realization.
2) Descriptive-style Directive: models receive more open-ended, qualitative style instructions expressed freely in natural language. It must parse this holistic description and infer the underlying parameter adjustments (e.g., prosody, speed, intensity) needed to produce the requested expressive style.
3) Role-play: models are given abstract, high-level role and scenario descriptions—closer to the kinds of prompts non-expert users might provide—and must leverage their own knowledge and contextual understanding to infer coherent acoustic expressions (emotion, volume, tone, etc.) and manifest these choices in the synthesized output.

To ensure realism, we build our dataset bottom-up: we mine highly expressive clips from movies and TV, apply rigorous cleaning and filtering, and then reverse-generate style instructions from the audio. In total, we offer three tasks with 1,000 English and 1,000 Chinese examples (6,000 instructions overall), each paired with a reference audio.
Finally, we leverage Gemini’s powerful speech-understanding capabilities—using an LLM-as-a-judge setup—to evaluate state-of-the-art instruct TTS systems. 
Results show that fine-grained acoustic control remains a major open challenge even for top-scoring models. Meanwhile, our case studies reveal the significant shortcomings in current TTS systems when it comes to reproducing natural vocal events, handling extreme emotional transitions, and synthesizing character-specific timbres—capabilities that are crucial for advancing TTS toward truly human-like expressiveness. 

In summary, our contributions are as follows:
\begin{itemize}
    \item We propose \textsc{InstructTTSEval}, an automatic benchmark for instruction-following TTS, comprising hierarchical tasks to comprehensively evaluate a model’s ability to interpret and execute complex natural-language style descriptions. 
    \item Confirming strong agreement with human annotations, we leverage Gemini as a judge to conduct rapid, scalable automatic assessment.
    \item We benchmark a diverse collection of popular open-source and commercial TTS systems on \textsc{InstructTTSEval}, providing detailed analysis for future model improvements.
\end{itemize}

\section{Related Work}


\begin{table*}[t]
    \centering
    \scalebox{0.87}{
    \begin{tabular}{l r l c c}
    \toprule
    \textbf{Dataset/Benchmark} &  \textbf{\# Labels} & \textbf{Highlights} & \textbf{Annotation}  & \textbf{Hier.} \\
    \midrule
    TextrolSpeech~\cite{Ji2023TextrolSpeechAT}      &  5  & - & Fixed  & \ding{55} \\
    PromptSpeech~\cite{Guo2022PromptttsCT} &  5 & - & Fixed   & \ding{55} \\
    SpeechCraft~\cite{Jin2024SpeechCraftAF} &8 & Emphasis & Fixed  & \ding{55} \\
    ParaSpeechCraft~\cite{Diwan2025ScalingRS} & 11 & Sound event & Fixed & \ding{55} \\
    \textbf{\textsc{InstructTTSEval}} \textbf{(ours)}  & 12  &Emphasis, sound event, dynamic changes  & Free-form  & \Checkmark \\
    \bottomrule
    \end{tabular}}
    \caption{Comparison with existing open-sourced style description datasets.  ``Hier.'' denotes hierarchical design. ``Fixed'' annotation refers to labels drawn from a limited set of tags or classifier outputs; ``free-form'' indicates natural-language descriptions that vary on a case-by-case basis. In all cases, the initial annotations are rewritten into fluent style instructions using a language model.}
    \label{tab:related}
\end{table*}

\subsection{Controllable TTS}
In previous work, researchers have taken several different approaches to controlling acoustic features in TTS.  CosyVoice~\cite{Du2024CosyVoiceAS} and FireRedTTS~\cite{Guo2024FireRedTTSAF} seek to insert special tokens into the input to steer the generated speech. ST-TTS~\cite{Kim2021ExpressiveTU} similarly uses dedicated style tags to modulate prosody and timbre.






Meanwhile, models such as InstructTTS~\cite{Yang2023InstructTTSME}, PromptTTS series~\cite{Guo2022PromptttsCT, Leng2023PromptTTS2D}, PromptStyle~\cite{Liu2023PromptStyleCS}, and ParlerTTS\cite{Lyth2024NaturalLG} allow users to describe desired voice characteristics in free-form text rather than relying on rigid tokens. 
CosyVoice 2~\cite{Du2024CosyVoice2S} also goes further by permitting a natural-language style prompt before the text. 
Salle~\cite{Ji2023TextrolSpeechAT} combines speech tags with LLM-based style rewriting; and ControlSpeech~\cite{Ji2024ControlSpeechTS} additionally integrates an example speech prompt for guidance. 
VoxInstruct~\cite{Zhou2024VoxInstructEH} merges style directives and transcript into a single instruction.
And EmoVoice~\cite{Yang2025EmoVoiceLE} focuses specifically on conveying emotional nuance via natural-language descriptions.
For commercial offerings, Services like Hume AI\footnote{ {} https://www.hume.ai/} and ElevenLabs\footnote{ {} https://elevenlabs.io/text-to-speech} let users simply type in style descriptions to shape output. The recent GPT-4o-mini TTS\footnote{ {} https://www.openai.fm/} and Gemini\footnote{ {} https://ai.google.dev/gemini-api/docs/speech-generation} likewise support free-form prompts for style control.
Despite this diversity of techniques, no unified evaluation framework exists to measure and compare their real-world style-control capabilities. To this end, we aim to propose a benchmark designed to assess how effectively these models follow user-defined style instructions.



\subsection{Acoustic-featured Datasets and Benchmarks}
%
\paragraph{Traditional TTS Evaluation Metrics}
Early TTS research has predominantly measured performance in terms of intelligibility and speaker similarity—most commonly using word error rate~(WER) and speaker-similarity~(SIM) scores~\cite{Panayotov2015LibrispeechAA, Anastassiou2024SeedTTSAF}. While these metrics effectively capture whether the synthesized speech is accurate and natural-sounding, they do not assess a system’s ability to follow detailed style or prosody instructions.
Meanwhile, subjective metrics such as Mean-Opinion-Score~(MOS) rely on human annotation, which is time-consuming and extremely costly.

\paragraph{Speech Understanding Benchmarks}
Benchmarks focusing on acoustic features mostly include speech understanding tasks. 
AudioBench~\cite{Wang2024AudioBenchAU} and SD-Eval~\cite{Ao2024SDEvalAB} integrate voice understanding tasks to assess models' ability to perceive paralinguistic information like accent, gender, and emotion. And Salmon \cite{Maimon2024ASF} measures whether the model can identify the inconsistencies in the input speech. These suites excel at evaluating recognition and classification, but they do not measure how well a TTS model can produce speech that matches a user-defined acoustic description.

\paragraph{Datasets with Style Description}
Several recent datasets pair speech samples with natural-language style descriptions using a variety of pipelines. TextrolSpeech~\cite{Ji2023TextrolSpeechAT} begins with five discrete features (gender, pitch, speed, volume, emotion) and relies on an LLM to weave those elements into coherent prompts. 
AudioBox~\cite{Vyas2023AudioboxUA} and PromptSpeech~\cite{Guo2022PromptttsCT} both have human annotators label clips along core dimensions and then apply an LLM to rewrite those tags into fluent sentences. And NLSpeech~\cite{Yang2023InstructTTSME} relies entirely on manually annotated data.
generates descriptive keywords (gender, speed, pitch, volume) before forming sentences and retrieving Spoken Language Understanding~(SLU)-tagged audio.
SpeechCraft~\cite{Jin2024SpeechCraftAF} adopts a bottom-up approach: a classifier first assigns tags to audio (e.g., ``elderly'', ``emphasis on...''), and an LLM then composes those tags into full descriptions. 
ParaSpeechCraft~\cite{Diwan2025ScalingRS} expands the tag set to 58 labels spanning inherent speaker traits and situational context. 

While each method offers useful insights, they face several challenges: 1) they cannot handle nuanced features in a speech, such as emotion transition; 2) they rely on heavy human annotation, or classifiers that are only available for limited features. We illustrate the major difference between ours and previous work in Tab.~\ref{tab:related}.
Therefore, we introduce an automatic and scalable pipeline for constructing style captions and instructions, capturing fine-grained acoustic attributes, to enable continuous evaluation and enhancement of controllable TTS systems.








\begin{figure*}[tbp]
    \centering
    \includegraphics[width=\linewidth]{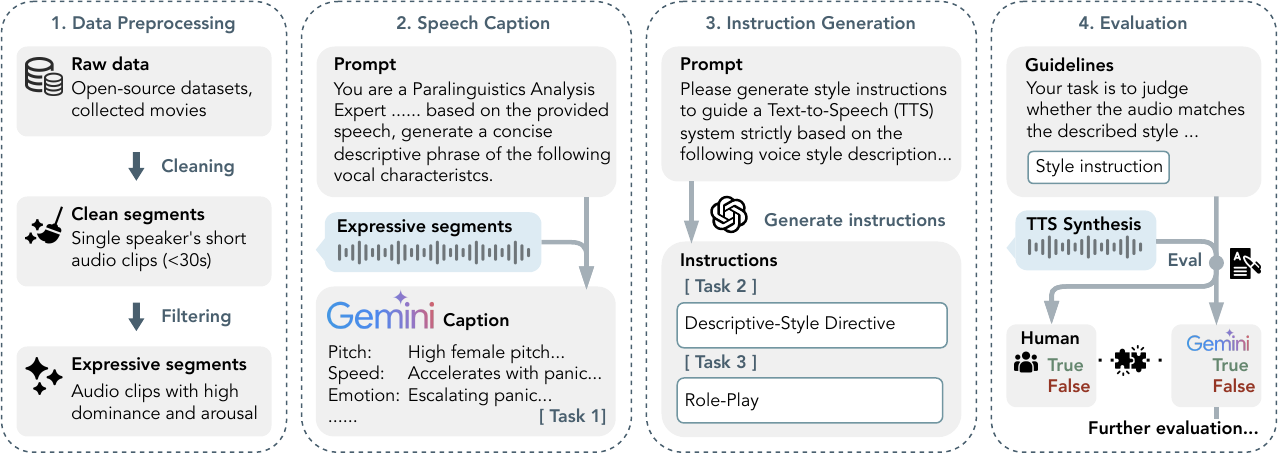}
    \caption{Overview of the benchmark construction and evaluation. We perform careful data cleaning and filtering to select audio with high expressiveness. Then we prompt Gemini to generate detailed, free-form descriptions for each acoustic feature; this caption is also used as the APS task's instruction. Then, we prompt GPT-4o to generate diverse instructions to create diverse DSD and RP instructions. Finally, after consistency verification between human and Gemini as judges, we further perform a holistic evaluation on current TTS systems. }
    \label{fig:pipeline}
\end{figure*}

\section{\textsc{InstructTTSEval}}

\subsection{Task Definition}
Based on previous studies~\cite{cutler1997prosody, Diwan2025ScalingRS, Jin2024SpeechCraftAF}, 
we integrate 12 features across four tiers: physiological (e.g., gender, pitch, texture), linguistic (e.g., clarity, fluency, speed), social (e.g., accent, age, volume), and psychological or pragmatic (e.g., emotion, tone, personality), as listed in Fig.~\ref{fig:gemeni_caption}. Building upon these features, we design the following tasks corresponding to three instruction granularities, as illustrated in Fig.~\ref{fig:tasks}, to evaluate controllable TTS systems:


\begin{enumerate}
    \item \textit{Acoustic-Parameter Specification}~(APS) focuses on fine-grained control over low-level acoustic attributes. The input consists of explicit instructions covering all 12 features, and the goal is to assess whether the model can independently manipulate each property. 
    \item \textit{Descriptive-Style Directive}~(DSD) is a more naturalistic variant where the structured instructions from APS are rewritten by an LLM into free-form descriptions. We further introduce diversity by randomly omitting some attributes in the prompt. This task examines the model’s ability to generalize from unstructured input and produce appropriate styles even when certain attributes are unspecified.
    \item \textit{Role-Play}~(RP) challenges the model’s contextual and social reasoning abilities. Instead of explicitly stating vocal traits, the prompts describe roles or scenarios (e.g., a teacher scolding a student, a nervous applicant in an interview). The model is expected to infer the corresponding vocal style based on world knowledge and map abstract social cues to concrete acoustic realizations, such as changes in speed, volume, intonation, or phrasing.
\end{enumerate}

Together, these three tasks offer a comprehensive benchmark for evaluating both low-level controllability and high-level style generalization in controllable TTS systems.

\subsection{Data Collection}
The overall process of how we construct \textsc{InstructTTSEval} is illustrated in Fig.~\ref{fig:pipeline}.

\subsubsection{Data Preprocessing}
\paragraph{Data Source}
To design prompts that exhibit strong stylistic expression with coherent feature alignment, we curate our data from movies, TV dramas, and variety shows—domains rich in expressive speech and diverse emotional delivery. We utilize NCSSD~\cite{Liu2024GenerativeEC} dataset and additionally collect and process supplementary audio from various film and television sources.


\paragraph{Data Cleaning} For our self-collected data, we apply speaker diarization~\cite{Bredin23, Plaquet23} to acquire segments shorter than 30 seconds of the same speakers. We then use \texttt{whisper-large-v3}~\cite{radford2022whisper} for automatic speech recognition~(ASR), followed by punctuation restoration using \texttt{ct-punc}~\footnote{ {} https://huggingface.co/funasr/ct-punc} and BELLE~\cite{BELLE}. This pipeline yields approximately 6,000 hours of transcribed speech.


\paragraph{Filtering} To ensure high audio quality, we filter both NCSSD and our own data using DNSMOS~\cite{reddy2021dnsmos} with a threshold of 2.8. We further refine the dataset with a custom-tuned WhisperD~\cite{darefsky2024parakeet} model to reserve single-speaker segments. To improve the accuracy of downstream speech caption, we retain only segments longer than 3 seconds and containing more than 10 words (for English) or 10 characters (for Chinese).
To select highly expressive speech samples, we employ the DVA toolkit~\cite{Wagner2022DawnOT}. We retain only the Dominance (Potency–Submissiveness) and Arousal (Activation–Deactivation) dimensions, filtering samples with both scores exceeding a threshold of 0.8. The Valence (V) dimension is discarded, as our benchmark aims to cover both positive and negative emotional expressions without bias toward emotional polarity.
Finally, we collect 2,000 segments as reference audio. Data statistics can be seen in Tab.~\ref{tab:Statistics}.

\begin{table}[htbp]
    \centering
    \begin{tabular}{l  rr  rr}
        \toprule
        \multirow{2}{*}{Source}  
        & \multicolumn{2}{c}{\# items} 
        & \multicolumn{2}{c}{Duration (h)} \\
        \cmidrule(lr){2-3} \cmidrule(lr){4-5}
        & EN & ZH & EN & ZH \\
        \midrule
        NCSSD       & 183 & 500       & 0.47  & 0.93  \\
        Collected   &  817 &  500      & 2.57  &  1.61   \\
        \midrule
        \textbf{Overall}   & 1,000  &  1,000     & 3.04 & 2.54\\
        \bottomrule
    \end{tabular}
    \caption{Statistics of reference audio.}
    \label{tab:Statistics}
\end{table}

\subsubsection{Speech Caption}
Compared to using predefined categorical tags, we argue that continuous natural language provides a more precise and nuanced description of the relevant speech features. To this end, we leverage the strong spoken language understanding~(SLU) capabilities of Gemini~\footnote{{} \texttt{gemini-2.5-pro-preview-05-06}} to generate a natural language description for each reference audio sample. The prompting strategy used for caption generation is illustrated in Fig.~\ref{fig:gemeni_caption}. This caption also serves as the instruction for the APS task.
In particular, we emphasize the importance of capturing \textit{dynamic changes} in vocal attributes. For example, a speaker might drop to a whisper when starting a gossip; or gradually escalate in emotional intensity, leading to increased volume and even shouting by the end. These dynamic transitions are common in natural speech but have often been overlooked in prior free-form style descriptions. We believe that modeling such temporal transitions is a crucial aspect of evaluating controllability in TTS systems.

\subsubsection{Instruction Generation}

Based on the generated captions, we use ChatGPT~\footnote{ {} \texttt{gpt-4o-2024-08-06}} to produce natural language style instructions.
For the Descriptive-Style Directive~(DSD) task, we randomly drop certain features from the instruction to simulate incomplete or underspecified input. The omitted features are considered unconstrained, allowing the TTS system to generate any plausible values for those attributes.
For the Role-play~(RP) task, we adopt a chain-of-thought (CoT) prompting strategy, guiding GPT to infer the speaker’s role or scenario by reasoning through the twelve defined acoustic features. This approach encourages the model to map low-level acoustic cues to high-level social or contextual roles.





\subsection{Metrics}
Inspired by \citet{Zhou2024VoxInstructEH}, we design an instruction-following metric to evaluate whether the synthesized speech aligns with the given instruction. A brief overview of the evaluation criteria is shown in Tab.~\ref{tab:scoring}, while detailed scoring guidelines are provided in App.~\ref{app:label}. For each subset, the final score is computed as the macro-average of the scores across all 1,000 items.

\begin{table}[htbp]
  \centering
  \begin{tabularx}{\columnwidth}{c X}
    \toprule
    \textbf{Score} & \textbf{Criteria} \\
    \midrule
    \textcolor[HTML]{324538}{\texttt{true}} &
      The sample’s primary style attributes (e.g., gender, pitch, rate, emotion)
      align with the prompt, without conflict. \\
    \addlinespace
    \textcolor[HTML]{943524}{\texttt{false}} &
      At least one key style attribute clearly conflicts with the prompt,
      or the overall style deviates from the prompt. \\
    \bottomrule
  \end{tabularx}
  \caption{Scoring Criteria.}
  \label{tab:scoring}
\end{table}
\section{Evaluation}

\subsection{Consistency}

To investigate whether Gemini can serve as a reliable substitute for human evaluation in our complex instruction-following tasks, we first assess the consistency between human judgments and Gemini's evaluations. We randomly selected 50 reference audio samples for human annotation across each of the three instruction types in both languages under investigation.
Among them, 25 samples are matched, meaning the instruction is originally derived from the corresponding reference audio. The remaining 25 are mismatched pairs, created by randomly assigning a non-corresponding instruction from the dataset to each reference audio. This setup allows us to assess Gemini's ability to correctly reject negative cases. Note, however, that due to the many-to-many~\cite{Ji2024ControlSpeechTS} nature of speech and description, it is still possible for a mismatched instruction to partially align with the given audio.

We recruit three human annotators~(graduate students, all with TOEFL scores above 100) and compensate them at a rate of 50 RMB/hour. They follow the same evaluation guidelines as Gemini. Detailed annotation guidance and screenshots are illustrated in App.~\ref{app:label}. We take the majority vote among the three annotators as the final human judgment, and compare it with Gemini's predictions. 
The agreement results are summarized in Tab.~\ref{tab:consistency}.
Notably, DSD and RP instructions are generated without feeding in the reference audio, so they may deviate from the original audio. Furthermore, because RP prompts are inherently more subjective, human–Gemini agreement tends to decline.

\begin{table}[tbp]
    \centering
    \begin{tabular}{ccccc}
    \toprule
    Accuracy & \textbf{APS} & \textbf{DSD} & \textbf{RP} & \textbf{Avg.} \\
    \midrule
       EN  & 86\% & 78\% & 66\% & 76.7\%\\
       ZH  & 88\% & 80\% & 76\% &  81.3\%\\
       \midrule
              \textbf{Avg.}  & 87\% & 79\% & 71\% & 79.0\% \\
    \bottomrule
    \end{tabular}
    \caption{Consistency between human and Gemini.}
    \label{tab:consistency}
\end{table}
 




\subsection{TTS Systems}
We select models that support free-style description as input, without requiring a prompt speech sample. For closed-sourced models, we evaluate gemini-2.5-flash-preview-tts~(gemini-flash), gemini-2.5-pro-preview-tts~(gemini-pro), gpt-4o-mini-tts, and Hume. ElevenLabs is only partially evaluated due to subscription limitations, so we exclude it from evaluation.
Since we must specify a voice for gemini-flash, gemini-pro, and gpt-4o-mini-tts, we randomly choose from the provided female/male voice list according to the caption result. For open-sourced TTS, we select Parler-TTS~\cite{Lyth2024NaturalLG}, VoxInstruct~\cite{Zhou2024VoxInstructEH}, and the reproducible variants~\cite{Ji2024ControlSpeechTS} of PromptTTS~\cite{Guo2022PromptttsCT} and PromptStyle~\cite{Liu2023PromptStyleCS} for evaluation. 
Each evaluation session containing all three tasks costs approximately \$12.8 for the EN-subset and \$11.9 for the ZH-subset using Gemini-as-a-judge.

\subsection{Results and Analysis}

\begin{table}[tbp]
    \centering
    \begin{tabular}{lrrrr}
        \toprule
        \textbf{TTS} & \textbf{APS} & \textbf{DSD} & \textbf{RP} & \textbf{Avg.} \\
        \midrule
        \rowcolor{mygray}
        reference\_audio  & 96.2 & 89.4 & 67.2 & 84.3\\
        \midrule
        \multicolumn{5}{>{\columncolor{myblue}}c}{\itshape Closed-sourced} \\
        \addlinespace
        gemini-flash* & 92.3 & 93.8 & 80.1 & 88.7 \\
        gemini-pro* & 87.6 & 86.0 & 67.2 & 80.3 \\
        gpt-4o-mini-tts   & 76.4 & 74.3 & 54.8 & 68.5 \\
        hume*             & 83.0 & 75.3 & 54.3 & 71.1 \\
        \midrule
        \multicolumn{5}{>{\columncolor{myblue}}c}{\itshape Open-sourced} \\
        \addlinespace
        VoxInstruct       & 54.9 & 57.0 & 39.3 & 50.4 \\
        Parler-TTS-mini   & 63.4 & 48.7 & 28.6 & 46.9 \\
        Parler-TTS-large  & 60.0 & 45.9 & 31.2 & 45.7 \\
        PromptTTS         & 64.3 & 47.2 & 31.4 & 47.6 \\
        PromptStyle       & 57.4 & 46.4 & 30.9 & 38.2 \\
        \bottomrule
    \end{tabular}
    \caption{Performance of TTS Systems on EN-subset.}
    \label{tab:performance-en}
\end{table}

\begin{table}[tbp]
    \centering
    \begin{tabular}{lrrrr}
        \toprule
        \textbf{TTS} & \textbf{APS} & \textbf{DSD} & \textbf{RP} & \textbf{Avg.} \\
        \midrule
        \rowcolor{mygray}
        reference\_audio  & 90.9 & 86.7 & 69.8 & 82.5 \\
        \midrule
        \multicolumn{5}{>{\columncolor{myblue}}c}{\itshape Closed-sourced} \\
        \addlinespace
        gemini-flash* & 88.2 & 90.9 & 77.3 & 85.4 \\
        gemini-pro* & 89.0 & 90.1 & 75.5 & 84.8 \\
        gpt-4o-mini-tts   & 54.9 & 52.3 & 46.0 & 51.1 \\
        \midrule
        \multicolumn{5}{>{\columncolor{myblue}}c}{\itshape Open-sourced} \\
        \addlinespace
        VoxInstruct      & 47.5 & 52.3 & 42.6 & 47.5 \\
        \bottomrule
    \end{tabular}
    \caption{Performance of TTS Systems on ZH-subset.}
    \label{tab:performance-zh}
\end{table}

\begin{table*}[t!]
  \centering
  \begin{tabularx}{\linewidth}{@{}p{0.64\linewidth} p{0.33\linewidth}@{}}
    \toprule
    \textbf{Instruction} & \textbf{Performance} \\
    \midrule
    \multicolumn{2}{>{\columncolor{myblue}}c}{\itshape Acoustic-Parameter Specification} \\
        \addlinespace
    Speed: Rapid pace initially, slightly slowing. \newline
    Volume: Energetic and relatively loud, decreasing slightly. \newline
    Emotion: \underline{Excited and emphatic}, ending with a \underline{sigh} suggesting weariness. 
    & \textbf{NO} existing models successfully `sigh'; gemini-flash, gemini-pro, hume, and gpt-4o-mini-tts sound excited. \\
    \midrule
    Volume: \underline{Shouting, very loud}. \newline
    Texture: Tense, somewhat strained. \newline
    Emotion: Intense \underline{panic and fear}. 
    &
    gpt-4o-mini-tts shows signs of `shouting'; hume bears anger. \\
    \midrule
    \multicolumn{2}{>{\columncolor{myblue}}c}{\itshape Descriptive-Style Directive} \\
        \addlinespace
    Begin with an artificially high-pitched, \underline{boisterous laugh} that carries a playful tone, then smoothly transition to a more deliberate pace with a standard conversational volume, subtly lowering the pitch afterward to deliver the rest with a slightly put-upon nasal quality. &
    gemini-flash, gemini-pro and gpt-4o-mini-tts successfully laugh out. \\
    \midrule
    Infuse your performance with an outgoing personality, ensuring a \underline{high child pitch} is woven through a swift, energetic delivery.
    &
    VoxInstruct and Parler-TTS-large generate child-like voices. \\
    \midrule
    \multicolumn{2}{>{\columncolor{myblue}}c}{\itshape Role-Play} \\
        \addlinespace
    Use an expressive and somewhat theatrical tone, like an \underline{elderly} British female storyteller sharing a funny story at a family gathering. Start with a quick pace and clear articulation, then \underline{slow down}, slightly fluctuating in pitch to emphasize key parts with a quirky and slightly bossy texture. &
    Parler-TTS-large generates a trembling voice of an elderly; VoxInstruct delivers a middle-aged to elderly female voice. \\
    \midrule
    Create an effect that keeps listeners focused: imagine a scenario where someone is \underline{shouting in panic and agony}, their words blurred and barely comprehensible, with \underline{piercing screams} that demand immediate attention. &
    \textbf{NO} models scream; gemini-flash, gemini-pro, and gpt-4o-mini-tts speaks as if it is in pain. \\
    \midrule
    Share the message with the energy of a young adult cartoon character, starting with a clear and calm voice that \underline{quickly rises} to a dynamic, emotionally impulsive pitch. &
    gemini-flash, gemini-pro, and VoxInstruct clearly raise voice; gpt-4o-mini-tts shows a slight rise. \\
    \midrule
    Infuse your tone with the brightness of a stage performer in a whimsical play, keeping your voice clear, lighthearted, and \underline{effortlessly melodic}. & 
    gemini-flash and gemini-pro successfully sing.
    \\
    \bottomrule
  \end{tabularx}
  \caption{Case studies. Models not mentioned indicate a lack of significant expressiveness. For APS instructions, some parameters are omitted due to length constraints.}
  \label{tab:case_study}
\end{table*}

We report the evaluation results in Tab.~\ref{tab:performance-en}~(EN-subset) and Tab.~\ref{tab:performance-zh}~(ZH-subset). * in tables denotes that few cases are missing due to model constraints such as max instruction length or safety settings.

Overall, closed-source commercial models significantly outperform open-source alternatives across all evaluation tasks, demonstrating a more precise control over acoustic information.
Notably, the Gemini series demonstrates exceptional performance both numerically and perceptually. However, the high scores may partly due to the tendency of LLMs to favor their own outputs~\cite{panickssery2024llmevaluatorsrecognizefavor}. 

A substantial gap exists between open-source and closed-source models. Among open-source alternatives, VoxInstruct significantly outperforms its peers on DSD and RP tasks, clearly standing out in naturalistic instruction interpretation. However, it struggles severely with APS tasks, likely due to its inability to process longer inputs, often leading to outputs that are undistinguishable. Parler-TTS-large and Parler-TTS-mini show no significant performance differences, with the mini version even outperforming the large variant on certain APS and DSD cases. PromptTTS and PromptStyle struggle most with generating expressive speech, yielding relatively flat and unremarkable samples that lack the dynamic vocal characteristics required for compelling synthesis.

It is worth mentioning that for the ZH subset, while closed-source models consistently outperform VoxInstruct in instruction-following capability, VoxInstruct's Chinese synthesis exhibits more native-speaker-like naturalness. This discrepancy likely stems from the predominantly English training data used by commercial models, resulting in less authentic Chinese prosodic patterns.

Despite the substantial progress demonstrated by commercial models, a significant gap in naturalness and expressiveness remains between all synthesized audio and reference audio. The best-performing systems still fall considerably short of highly controllable, natural human-level synthesis, indicating that fine-grained paralinguistic control represents a major open challenge for the field. 

\subsection{Case Studies}
In this section, we select some representative cases for analysis~(Tab.~\ref{tab:case_study}). 

First, \textbf{modern TTS models still struggle to reproduce the paralinguistic \textit{sound events} that frequently occur in human speech}, such as sighs, sudden bursts of laughter, screams, etc. In our case study, only gpt-4o-mini-tts, gemini-flash, and gemini-pro are able to generate laughter. Yet these vocal events are essential for conveying emotion and maintaining a natural speech flow; a powerful, controllable TTS system should be able to capture and synthesize them.


\textbf{Few models are capable of extreme emotional expressions and rapid affective shifts.} In our evaluation, gpt-4o-mini-tts clearly stands out: it can produce controlled shouting and other heightened vocalizations on demand. 
For dynamic emotional transitions, gemini-flash, gemini-pro, and VoxInstruct demonstrate initial capability of handling some transitions, such as voice rise; but still struggle to handle emotion change (here from calm to impulsive). 



Interestingly, in certain timbre-focused cases, \textbf{some open-sourced systems actually can produce surprisingly impressive results}. Take the ``child voice'' scenario, for example: hume is blocked from generating child-like timbres by its safety filters; gemini-flash, gemini-pro, and gpt-4o-mini-tts cannot stray from its fixed voice settings, which largely limit its prosodic generality. Yet several open-source TTS models—despite weaker overall emotional control—deliver astonishingly authentic, youthful vocal qualities. And Parler-TTS-Large successfully synthesizes the voice of the elderly. This suggests that timbre flexibility and emotional expressiveness remain orthogonal capabilities, and that future TTS research should aim to unify them rather than treat them separately.

Finally, \textbf{singing remains beyond current capabilities}. Only gemini-flash and gemini-pro show some capability for singing effects when instructed to infuse melodic tone. Instructing a TTS model to “speak as if singing” requires integrating prosody, melody, emotion, timbre, and rhythm in a cohesive way. This multidimensional coordination sets a much higher bar for naturalness and expressiveness in controllable TTS—and represents a key direction for future research.

\section{Conclusions}





In this paper, we carefully construct a hirerachical benchmark for instruction-following TTS. We meticulously design a three-tier evaluation task—spanning the low-level Acoustic-Parameter Specification~(APS) task, the mid-level Descriptive-Style Directive~(DSD) task, and the high-level Role-Play~(RP) task—to comprehensively measure current TTS systems’ ability to follow complex natural-language descriptions of acoustic features. 
Our results reveal that existing models still struggle with fine-grained paralinguistic control, and expose significant performance gaps both between closed- and open-source systems and across different languages. Moreover, our case studies highlight major deficiencies in reproducing natural vocal events, managing extreme emotional transitions, and synthesizing character-specific timbres—capabilities that are crucial for advancing TTS toward truly human-like expressiveness. We hope this benchmark will catalyze further progress in developing more controllable and expressive speech synthesis.

\clearpage
\newpage
\section*{Limitations}


We acknowledge that our work may have the following limitations: 1) Subjectivity and evaluation cost. Some of our tasks, particularly for the Role-Play (RP), are inherently subjective. Inter-annotator agreement among human raters is relatively low compared to APS and DSD tasks, which introduces noise when using automated evaluators like Gemini. Moreover, continuously conducting large-scale evaluations using Gemini is cost-intensive. In future work, we plan to develop a more accurate and cost-efficient evaluator to enact iterative evaluation.
2) Data imbalance. Since our dataset is constructed in a bottom-up fashion from specific acoustic events and style directives, certain classes (e.g., particular emotions or role archetypes) are underrepresented. This imbalance could bias model performance and evaluation. We will expand the benchmark to include a broader, more evenly distributed set of style categories.
\section*{Ethical Considerations}

Owing to our large-scale automated pipeline, we are unable to manually review every text–instruction pair. As a result, it may consist of some inappropriate content. Please note that any content from the reference audio and the synthesized audio does NOT reflect the authors’ views or endorsements. Additionally, this dataset and benchmark are intended solely for academic and research use.


\bibliography{acl_latex}

\begin{thebibliography}{39}
\providecommand{\natexlab}[1]{#1}

\bibitem[{Anastassiou et~al.(2024)Anastassiou, Chen, Chen, Chen, Chen, Chen, Cong, Deng, Ding, Gao, Gong, Huang, Huang, Huang, Huo, Jia, Li, Li, Li, Li, Li, Li, Liu, Liu, Liu, Liu, Liu, Liu, Lu, Pan, Wang, Wang, Wang, Wei, Wu, Yao, Yang, Yi, Zhang, Zhang, Zhang, Zhang, Zhang, Zhao, Zhong, and Zhuang}]{Anastassiou2024SeedTTSAF}
Philip Anastassiou, Jiawei Chen, Jitong Chen, Yuanzhe Chen, Zhuo Chen, Ziyi Chen, Jian Cong, Lelai Deng, Chuang Ding, Lu~Gao, Mingqing Gong, Peisong Huang, Qingqing Huang, Zhiying Huang, Yuanyuan Huo, Dongya Jia, Chumin Li, Feiya Li, Hui Li, and 27 others. 2024.
\newblock \href {https://api.semanticscholar.org/CorpusID:270226353} {Seed-tts: A family of high-quality versatile speech generation models}.
\newblock \emph{ArXiv}, abs/2406.02430.

\bibitem[{Ao et~al.(2024)Ao, Wang, Tian, Chen, Zhang, Lu, Wang, Li, and Wu}]{Ao2024SDEvalAB}
Junyi Ao, Yuancheng Wang, Xiaohai Tian, Dekun Chen, Jun Zhang, Lu~Lu, Yuxuan Wang, Haizhou Li, and Zhizheng Wu. 2024.
\newblock \href {https://api.semanticscholar.org/CorpusID:270620454} {Sd-eval: A benchmark dataset for spoken dialogue understanding beyond words}.
\newblock \emph{ArXiv}, abs/2406.13340.

\bibitem[{BELLEGroup(2023)}]{BELLE}
BELLEGroup. 2023.
\newblock Belle: Be everyone's large language model engine.
\newblock \url{https://github.com/LianjiaTech/BELLE}.

\bibitem[{Betker(2023)}]{Betker2023BetterSS}
James Betker. 2023.
\newblock \href {https://api.semanticscholar.org/CorpusID:258676394} {Better speech synthesis through scaling}.
\newblock \emph{ArXiv}, abs/2305.07243.

\bibitem[{Bredin(2023)}]{Bredin23}
Hervé Bredin. 2023.
\newblock {pyannote.audio 2.1 speaker diarization pipeline: principle, benchmark, and recipe}.
\newblock In \emph{Proc. INTERSPEECH 2023}.

\bibitem[{Chen et~al.(2024)Chen, Niu, Ma, Deng, Wang, Zhao, Yu, and Chen}]{Chen2024F5TTSAF}
Yushen Chen, Zhikang Niu, Ziyang Ma, Keqi Deng, Chunhui Wang, Jian Zhao, Kai Yu, and Xie Chen. 2024.
\newblock \href {https://api.semanticscholar.org/CorpusID:273228169} {F5-tts: A fairytaler that fakes fluent and faithful speech with flow matching}.
\newblock \emph{ArXiv}, abs/2410.06885.

\bibitem[{Cutler et~al.(1997)Cutler, Dahan, and Van~Donselaar}]{cutler1997prosody}
Anne Cutler, Delphine Dahan, and Wilma Van~Donselaar. 1997.
\newblock Prosody in the comprehension of spoken language: A literature review.
\newblock \emph{Language and speech}, 40(2):141--201.

\bibitem[{Darefsky et~al.(2024)Darefsky, Zhu, and Duan}]{darefsky2024parakeet}
Jordan Darefsky, Ge~Zhu, and Zhiyao Duan. 2024.
\newblock \href {https://jordandarefsky.com/blog/2024/parakeet/} {Parakeet}.

\bibitem[{Diwan et~al.(2025)Diwan, Zheng, Harwath, and Choi}]{Diwan2025ScalingRS}
Anuj Diwan, Zhisheng Zheng, David Harwath, and Eunsol Choi. 2025.
\newblock \href {https://api.semanticscholar.org/CorpusID:276813138} {Scaling rich style-prompted text-to-speech datasets}.
\newblock \emph{ArXiv}, abs/2503.04713.

\bibitem[{Du et~al.(2024{\natexlab{a}})Du, Chen, Zhang, Hu, Lu, Yang, Hu, Zheng, Gu, Ma, Gao, and Yan}]{Du2024CosyVoiceAS}
Zhihao Du, Qian Chen, Shiliang Zhang, Kai Hu, Heng Lu, Yexin Yang, Hangrui Hu, Siqi Zheng, Yue Gu, Ziyang Ma, Zhifu Gao, and Zhijie Yan. 2024{\natexlab{a}}.
\newblock \href {https://api.semanticscholar.org/CorpusID:271050381} {Cosyvoice: A scalable multilingual zero-shot text-to-speech synthesizer based on supervised semantic tokens}.
\newblock \emph{ArXiv}, abs/2407.05407.

\bibitem[{Du et~al.(2024{\natexlab{b}})Du, Wang, Chen, Shi, Lv, Zhao, Gao, Yang, Gao, Wang, Yu, Liu, Sheng, Gu, Deng, Wang, Zhang, Yan, and Zhou}]{Du2024CosyVoice2S}
Zhihao Du, Yuxuan Wang, Qian Chen, Xian Shi, Xiang Lv, Tianyu Zhao, Zhifu Gao, Yexin Yang, Changfeng Gao, Hui Wang, Fan Yu, Huadai Liu, Zhengyan Sheng, Yue Gu, Chong Deng, Wen Wang, Shiliang Zhang, Zhijie Yan, and Jing-Ru Zhou. 2024{\natexlab{b}}.
\newblock \href {https://api.semanticscholar.org/CorpusID:274762932} {Cosyvoice 2: Scalable streaming speech synthesis with large language models}.
\newblock \emph{ArXiv}, abs/2412.10117.

\bibitem[{Guo et~al.(2024)Guo, Liu, Shen, Wu, Xie, Xie, and Xu}]{Guo2024FireRedTTSAF}
Hao-Han Guo, Kun Liu, Fei-Yu Shen, Yi-Chen Wu, Fenglong Xie, Kun Xie, and Kai-Tuo Xu. 2024.
\newblock \href {https://api.semanticscholar.org/CorpusID:272424176} {Fireredtts: A foundation text-to-speech framework for industry-level generative speech applications}.
\newblock \emph{ArXiv}, abs/2409.03283.

\bibitem[{Guo et~al.(2022)Guo, Leng, Wu, Zhao, and Tan}]{Guo2022PromptttsCT}
Zhifang Guo, Yichong Leng, Yihan Wu, Sheng Zhao, and Xuejiao Tan. 2022.
\newblock \href {https://api.semanticscholar.org/CorpusID:253761189} {Prompttts: Controllable text-to-speech with text descriptions}.
\newblock \emph{ICASSP 2023 - 2023 IEEE International Conference on Acoustics, Speech and Signal Processing (ICASSP)}, pages 1--5.

\bibitem[{Ji et~al.(2023)Ji, Zuo, Fang, Jiang, Chen, Duan, Huai, and Zhao}]{Ji2023TextrolSpeechAT}
Shengpeng Ji, Jialong Zuo, Minghui Fang, Ziyue Jiang, Feiyang Chen, Xinyu Duan, Baoxing Huai, and Zhou Zhao. 2023.
\newblock \href {https://api.semanticscholar.org/CorpusID:261242529} {Textrolspeech: A text style control speech corpus with codec language text-to-speech models}.
\newblock \emph{ICASSP 2024 - 2024 IEEE International Conference on Acoustics, Speech and Signal Processing (ICASSP)}, pages 10301--10305.

\bibitem[{Ji et~al.(2024)Ji, Zuo, Fang, Zheng, Chen, Wang, Jiang, Huang, Cheng, Huang, and Zhao}]{Ji2024ControlSpeechTS}
Shengpeng Ji, Jialong Zuo, Minghui Fang, Siqi Zheng, Qian Chen, Wen Wang, Ziyue Jiang, Hai Huang, Xize Cheng, Rongjie Huang, and Zhou Zhao. 2024.
\newblock \href {https://api.semanticscholar.org/CorpusID:270216344} {Controlspeech: Towards simultaneous zero-shot speaker cloning and zero-shot language style control with decoupled codec}.
\newblock \emph{ArXiv}, abs/2406.01205.

\bibitem[{Jin et~al.(2024)Jin, Jia, Wang, Li, Zhou, Zhou, Qin, and Wu}]{Jin2024SpeechCraftAF}
Zeyu Jin, Jia Jia, Qixin Wang, Kehan Li, Shuoyi Zhou, Songtao Zhou, Xiaoyu Qin, and Zhiyong Wu. 2024.
\newblock \href {https://api.semanticscholar.org/CorpusID:271956527} {Speechcraft: A fine-grained expressive speech dataset with natural language description}.
\newblock \emph{Proceedings of the 32nd ACM International Conference on Multimedia}.

\bibitem[{Kim et~al.(2021)Kim, Cheon, Choi, Kim, and Kim}]{Kim2021ExpressiveTU}
Minchan Kim, Sung~Jun Cheon, Byoung~Jin Choi, Jong~Jin Kim, and Nam~Soo Kim. 2021.
\newblock \href {https://api.semanticscholar.org/CorpusID:232478871} {Expressive text-to-speech using style tag}.
\newblock In \emph{Interspeech}.

\bibitem[{Leng et~al.(2023)Leng, Guo, Shen, Tan, Ju, Liu, Liu, Yang, Zhang, Song, He, Li, Zhao, Qin, and Bian}]{Leng2023PromptTTS2D}
Yichong Leng, Zhifang Guo, Kai Shen, Xu~Tan, Zeqian Ju, Yanqing Liu, Yufei Liu, Dongchao Yang, Leying Zhang, Kaitao Song, Lei He, Xiang-Yang Li, Sheng Zhao, Tao Qin, and Jiang Bian. 2023.
\newblock \href {https://api.semanticscholar.org/CorpusID:261557296} {Prompttts 2: Describing and generating voices with text prompt}.
\newblock \emph{ArXiv}, abs/2309.02285.

\bibitem[{Liao et~al.(2024)Liao, Wang, Li, Cheng, Zhang, Zhou, and Xing}]{fish-speech-v1.4}
Shijia Liao, Yuxuan Wang, Tianyu Li, Yifan Cheng, Ruoyi Zhang, Rongzhi Zhou, and Yijin Xing. 2024.
\newblock \href {https://arxiv.org/abs/2411.01156} {Fish-speech: Leveraging large language models for advanced multilingual text-to-speech synthesis}.
\newblock \emph{Preprint}, arXiv:2411.01156.

\bibitem[{Liu et~al.(2023)Liu, Zhang, Lei, Chen, Wang, Li, and Xie}]{Liu2023PromptStyleCS}
Guanghou Liu, Yongmao Zhang, Yinjiao Lei, Yunlin Chen, Rui Wang, Zhifei Li, and Linfu Xie. 2023.
\newblock \href {https://api.semanticscholar.org/CorpusID:258987676} {Promptstyle: Controllable style transfer for text-to-speech with natural language descriptions}.
\newblock In \emph{Interspeech}.

\bibitem[{Liu et~al.(2024)Liu, Hu, Ren, Yin, and Li}]{Liu2024GenerativeEC}
Rui Liu, Yifan Hu, Yi~Ren, Xiang Yin, and Haizhou Li. 2024.
\newblock \href {https://api.semanticscholar.org/CorpusID:271571519} {Generative expressive conversational speech synthesis}.
\newblock \emph{Proceedings of the 32nd ACM International Conference on Multimedia}.

\bibitem[{Lyth and King(2024)}]{Lyth2024NaturalLG}
Daniel Lyth and Simon King. 2024.
\newblock \href {https://api.semanticscholar.org/CorpusID:267412845} {Natural language guidance of high-fidelity text-to-speech with synthetic annotations}.
\newblock \emph{ArXiv}, abs/2402.01912.

\bibitem[{Maimon et~al.(2024)Maimon, Roth, and Adi}]{Maimon2024ASF}
Gallil Maimon, Amit Roth, and Yossi Adi. 2024.
\newblock \href {https://api.semanticscholar.org/CorpusID:272593001} {A suite for acoustic language model evaluation}.
\newblock \emph{ArXiv}, abs/2409.07437.

\bibitem[{Panayotov et~al.(2015)Panayotov, Chen, Povey, and Khudanpur}]{Panayotov2015LibrispeechAA}
Vassil Panayotov, Guoguo Chen, Daniel Povey, and Sanjeev Khudanpur. 2015.
\newblock \href {https://api.semanticscholar.org/CorpusID:2191379} {Librispeech: An asr corpus based on public domain audio books}.
\newblock \emph{2015 IEEE International Conference on Acoustics, Speech and Signal Processing (ICASSP)}, pages 5206--5210.

\bibitem[{Panickssery et~al.(2024)Panickssery, Bowman, and Feng}]{panickssery2024llmevaluatorsrecognizefavor}
Arjun Panickssery, Samuel~R. Bowman, and Shi Feng. 2024.
\newblock \href {https://arxiv.org/abs/2404.13076} {Llm evaluators recognize and favor their own generations}.
\newblock \emph{Preprint}, arXiv:2404.13076.

\bibitem[{Plaquet and Bredin(2023)}]{Plaquet23}
Alexis Plaquet and Hervé Bredin. 2023.
\newblock {Powerset multi-class cross entropy loss for neural speaker diarization}.
\newblock In \emph{Proc. INTERSPEECH 2023}.

\bibitem[{Radford et~al.(2022)Radford, Kim, Xu, Brockman, McLeavey, and Sutskever}]{radford2022whisper}
Alec Radford, Jong~Wook Kim, Tao Xu, Greg Brockman, Christine McLeavey, and Ilya Sutskever. 2022.
\newblock \href {https://doi.org/10.48550/ARXIV.2212.04356} {Robust speech recognition via large-scale weak supervision}.
\newblock \emph{arXiv preprint}.

\bibitem[{Reddy et~al.(2021)Reddy, Gopal, and Cutler}]{reddy2021dnsmos}
Chandan~KA Reddy, Vishak Gopal, and Ross Cutler. 2021.
\newblock Dnsmos: A non-intrusive perceptual objective speech quality metric to evaluate noise suppressors.
\newblock In \emph{ICASSP 2021 IEEE International Conference on Acoustics, Speech and Signal Processing (ICASSP)}, pages 6493--6497. IEEE.

\bibitem[{Ren et~al.(2020)Ren, Hu, Tan, Qin, Zhao, Zhao, and Liu}]{Ren2020FastSpeech2F}
Yi~Ren, Chenxu Hu, Xu~Tan, Tao Qin, Sheng Zhao, Zhou Zhao, and Tie-Yan Liu. 2020.
\newblock \href {https://api.semanticscholar.org/CorpusID:219531522} {Fastspeech 2: Fast and high-quality end-to-end text to speech}.
\newblock \emph{ArXiv}, abs/2006.04558.

\bibitem[{Vyas et~al.(2023)Vyas, Shi, Le, Tjandra, Wu, Guo, Zhang, Zhang, Adkins, Ngan, Wang, Cruz, Akula, Akinyemi, Ellis, Moritz, Yungster, Rakotoarison, Tan, Summers, Wood, Lane, Williamson, and Hsu}]{Vyas2023AudioboxUA}
Apoorv Vyas, Bowen Shi, Matt Le, Andros Tjandra, Yi-Chiao Wu, Baishan Guo, Jiemin Zhang, Xinyue Zhang, Robert Adkins, W.K.F. Ngan, Jeff Wang, Ivan Cruz, Bapi Akula, Akinniyi~Tunde Akinyemi, Brian Ellis, Rashel Moritz, Yael Yungster, Alice Rakotoarison, Liang Tan, and 5 others. 2023.
\newblock \href {https://api.semanticscholar.org/CorpusID:266551778} {Audiobox: Unified audio generation with natural language prompts}.
\newblock \emph{ArXiv}, abs/2312.15821.

\bibitem[{Wagner et~al.(2022)Wagner, Triantafyllopoulos, Wierstorf, Schmitt, Eyben, and Schuller}]{Wagner2022DawnOT}
Johannes Wagner, Andreas Triantafyllopoulos, Hagen Wierstorf, Maximilian Schmitt, Florian Eyben, and Bj{\"o}rn Schuller. 2022.
\newblock \href {https://api.semanticscholar.org/CorpusID:247451056} {Dawn of the transformer era in speech emotion recognition: Closing the valence gap}.
\newblock \emph{IEEE Transactions on Pattern Analysis and Machine Intelligence}, 45:10745--10759.

\bibitem[{Wang et~al.(2024{\natexlab{a}})Wang, Zou, Lin, Sun, Liu, Zhang, Liu, Aw, and Chen}]{Wang2024AudioBenchAU}
Bin Wang, Xunlong Zou, Geyu Lin, Shuo Sun, Zhuohan Liu, Wenyu Zhang, Zhengyuan Liu, AiTi Aw, and Nancy~F. Chen. 2024{\natexlab{a}}.
\newblock \href {https://api.semanticscholar.org/CorpusID:270702586} {Audiobench: A universal benchmark for audio large language models}.
\newblock \emph{ArXiv}, abs/2406.16020.

\bibitem[{Wang et~al.(2023)Wang, Chen, Wu, Zhang, Zhou, Liu, Chen, Liu, Wang, Li, He, Zhao, and Wei}]{Wang2023NeuralCL}
Chengyi Wang, Sanyuan Chen, Yu~Wu, Zi-Hua Zhang, Long Zhou, Shujie Liu, Zhuo Chen, Yanqing Liu, Huaming Wang, Jinyu Li, Lei He, Sheng Zhao, and Furu Wei. 2023.
\newblock \href {https://api.semanticscholar.org/CorpusID:255440307} {Neural codec language models are zero-shot text to speech synthesizers}.
\newblock \emph{IEEE Transactions on Audio, Speech and Language Processing}, 33:705--718.

\bibitem[{Wang et~al.(2025)Wang, Jiang, Ma, Zhang, Liu, Li, Liang, Zheng, Wang, Feng, Bian, Ye, Cheng, Yuan, Zhao, Zhu, Pan, Xue, Zhu, Chen, Li, Chen, Xie, Guo, and Xue}]{wang2025sparktts}
Xinsheng Wang, Mingqi Jiang, Ziyang Ma, Ziyu Zhang, Songxiang Liu, Linqin Li, Zheng Liang, Qixi Zheng, Rui Wang, Xiaoqin Feng, Weizhen Bian, Zhen Ye, Sitong Cheng, Ruibin Yuan, Zhixian Zhao, Xinfa Zhu, Jiahao Pan, Liumeng Xue, Pengcheng Zhu, and 6 others. 2025.
\newblock \href {https://arxiv.org/abs/2503.01710} {Spark-tts: An efficient llm-based text-to-speech model with single-stream decoupled speech tokens}.
\newblock \emph{Preprint}, arXiv:2503.01710.

\bibitem[{Wang et~al.(2024{\natexlab{b}})Wang, Zhan, Liu, Zeng, Guo, Zheng, Zhang, Zhang, Zhang, and Wu}]{Wang2024MaskGCTZT}
Yuancheng Wang, Haoyue Zhan, Liwei Liu, Ruihong Zeng, Haotian Guo, Jiachen Zheng, Qiang Zhang, Xueyao Zhang, Shunsi Zhang, and Zhizheng Wu. 2024{\natexlab{b}}.
\newblock \href {https://api.semanticscholar.org/CorpusID:272366599} {Maskgct: Zero-shot text-to-speech with masked generative codec transformer}.
\newblock \emph{ArXiv}, abs/2409.00750.

\bibitem[{Yang et~al.(2023)Yang, Liu, Huang, Lei, Weng, Meng, and Yu}]{Yang2023InstructTTSME}
Dongchao Yang, Songxiang Liu, Rongjie Huang, Guangzhi Lei, Chao Weng, Helen~M. Meng, and Dong Yu. 2023.
\newblock \href {https://api.semanticscholar.org/CorpusID:256416291} {Instructtts: Modelling expressive tts in discrete latent space with natural language style prompt}.
\newblock \emph{IEEE/ACM Transactions on Audio, Speech, and Language Processing}, 32:2913--2925.

\bibitem[{Yang et~al.(2025)Yang, Yang, Chen, Ma, Chen, Wang, Wang, Yang, Niu, Liu, Yu, Du, Gao, Zhang, and Chen}]{Yang2025EmoVoiceLE}
Guanrou Yang, Chen Yang, Qian Chen, Ziyang Ma, Wenxi Chen, Wen Wang, Tianrui Wang, Yifan Yang, Zhikang Niu, Wenrui Liu, Fan Yu, Zhihao Du, Zhifu Gao, Shiliang Zhang, and Xie Chen. 2025.
\newblock \href {https://api.semanticscholar.org/CorpusID:277857208} {Emovoice: Llm-based emotional text-to-speech model with freestyle text prompting}.

\bibitem[{Ye et~al.(2025)Ye, Zhu, min Chan, Wang, Tan, Lei, Peng, Liu, Jin, Dai, Lin, Chen, Du, Xue, Chen, Li, Xie, Kong, Guo, and Xue}]{Ye2025LlasaST}
Zhen Ye, Xinfa Zhu, Chi min Chan, Xinsheng Wang, Xu~Tan, Jiahe Lei, Yi~Peng, Haohe Liu, Yizhu Jin, Zheqi Dai, Hongzhan Lin, Jianyi Chen, Xingjian Du, Liumeng Xue, Yunlin Chen, Zhifei Li, Lei Xie, Qiuqiang Kong, Yi-Ting Guo, and Wei Xue. 2025.
\newblock \href {https://api.semanticscholar.org/CorpusID:276161207} {Llasa: Scaling train-time and inference-time compute for llama-based speech synthesis}.
\newblock \emph{ArXiv}, abs/2502.04128.

\bibitem[{Zhou et~al.(2024)Zhou, Qin, Jin, Zhou, Lei, Zhou, Wu, and Jia}]{Zhou2024VoxInstructEH}
Yixuan Zhou, Xiaoyu Qin, Zeyu Jin, Shuoyi Zhou, Shunwei Lei, Songtao Zhou, Zhiyong Wu, and Jia Jia. 2024.
\newblock \href {https://api.semanticscholar.org/CorpusID:271974317} {Voxinstruct: Expressive human instruction-to-speech generation with unified multilingual codec language modelling}.
\newblock \emph{Proceedings of the 32nd ACM International Conference on Multimedia}.

\end{thebibliography}

\clearpage
\newpage
\appendix
\section{Speech Caption}
\begin{figure*}
    \centering
    \includegraphics[width=\linewidth]{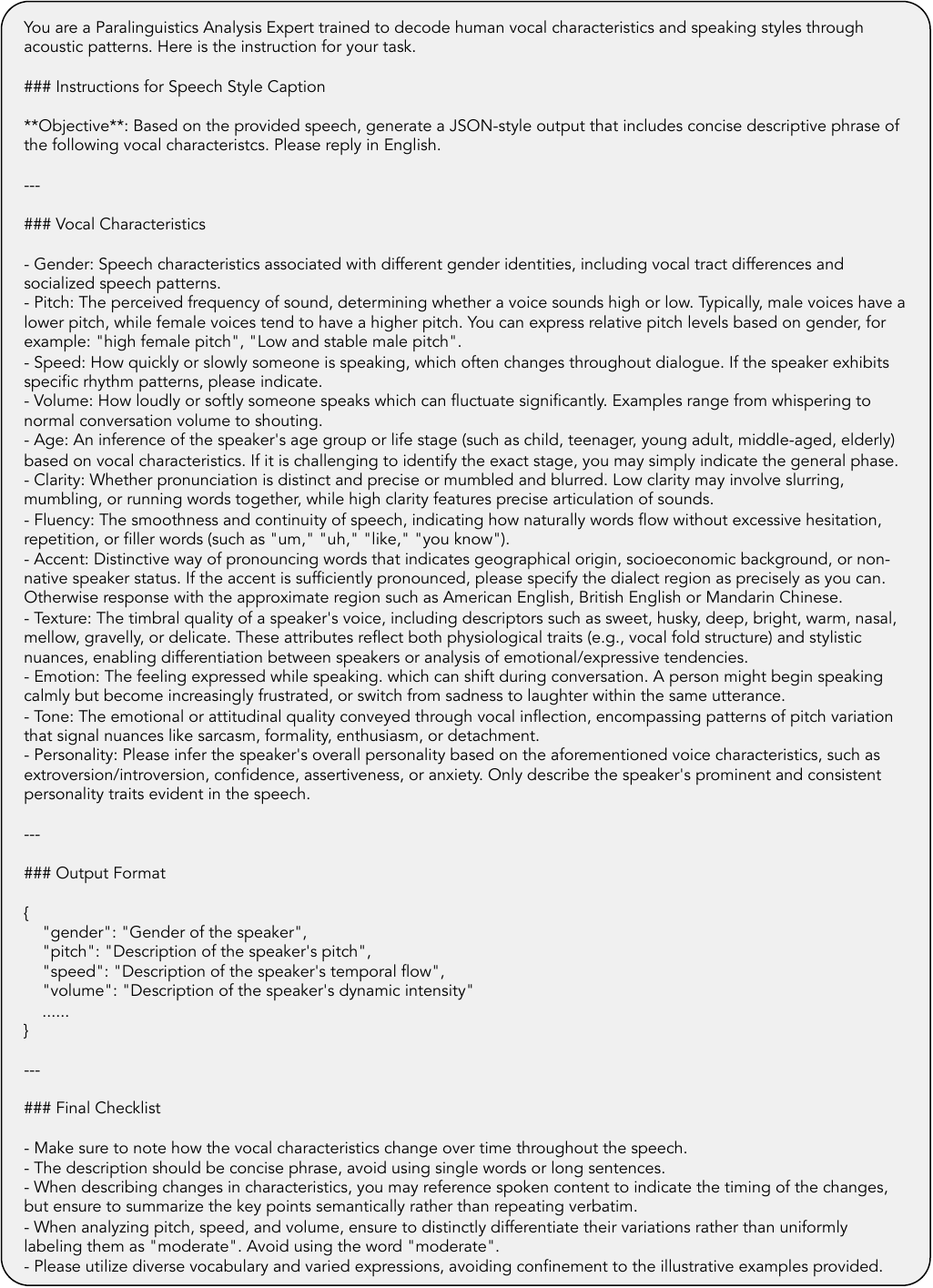}
    \caption{Speech caption prompt for Gemini.}
    \label{fig:gemeni_caption}
\end{figure*}

Fig.~\ref{fig:gemeni_caption} illustrates our prompt for Gemini to generate a detailed, natural language description for each speech segments.


\section{Instruction Generation}
In this section, we introduce our prompting strategies for generating style instructions. In order to increase diversity and accuracy, we employ: 1) random dropouts, 2) multiple prompts, 3) generate more instructions at one time, and 4) utilize the Chain-of-Thought~(CoT) strategy. 

\begin{figure*}
    \centering
    \includegraphics[width=\linewidth]{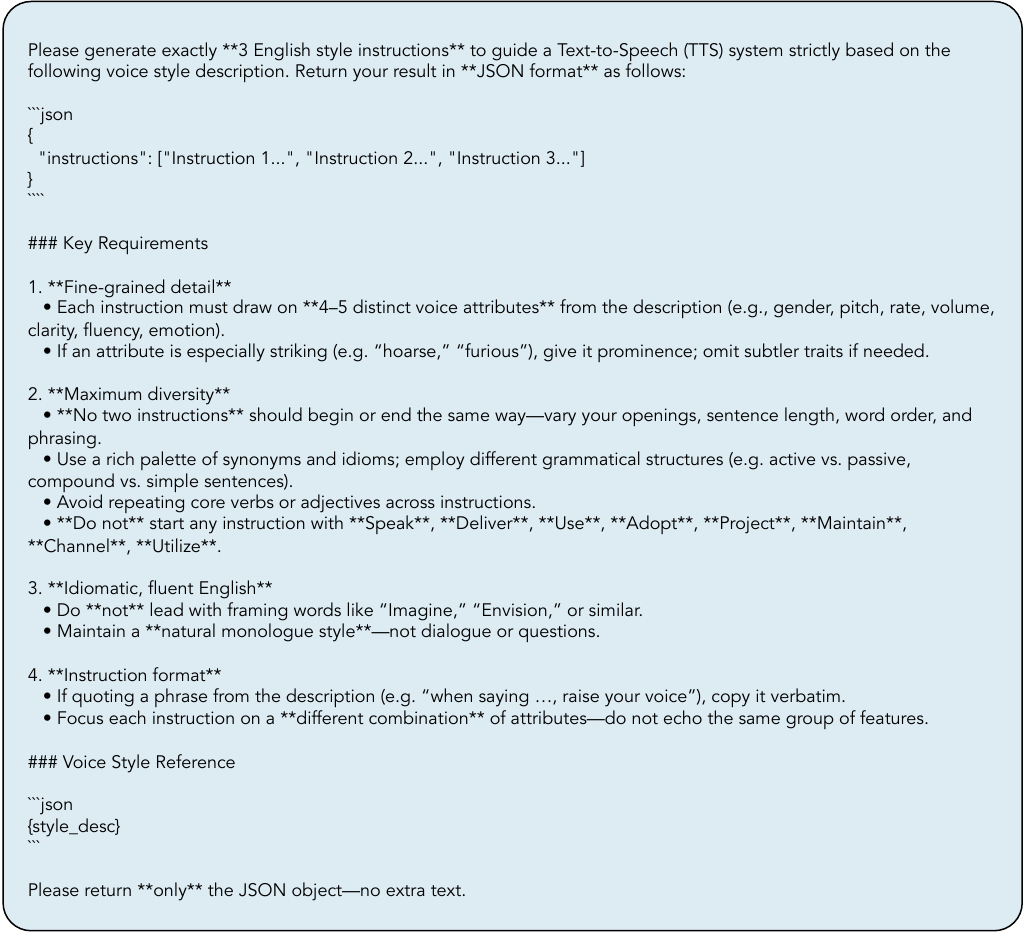}
    \caption{Prompts for DSD instruction generation.}
    \label{fig:dsd}
\end{figure*}

\begin{figure*}
    \centering
    \includegraphics[width=\linewidth]{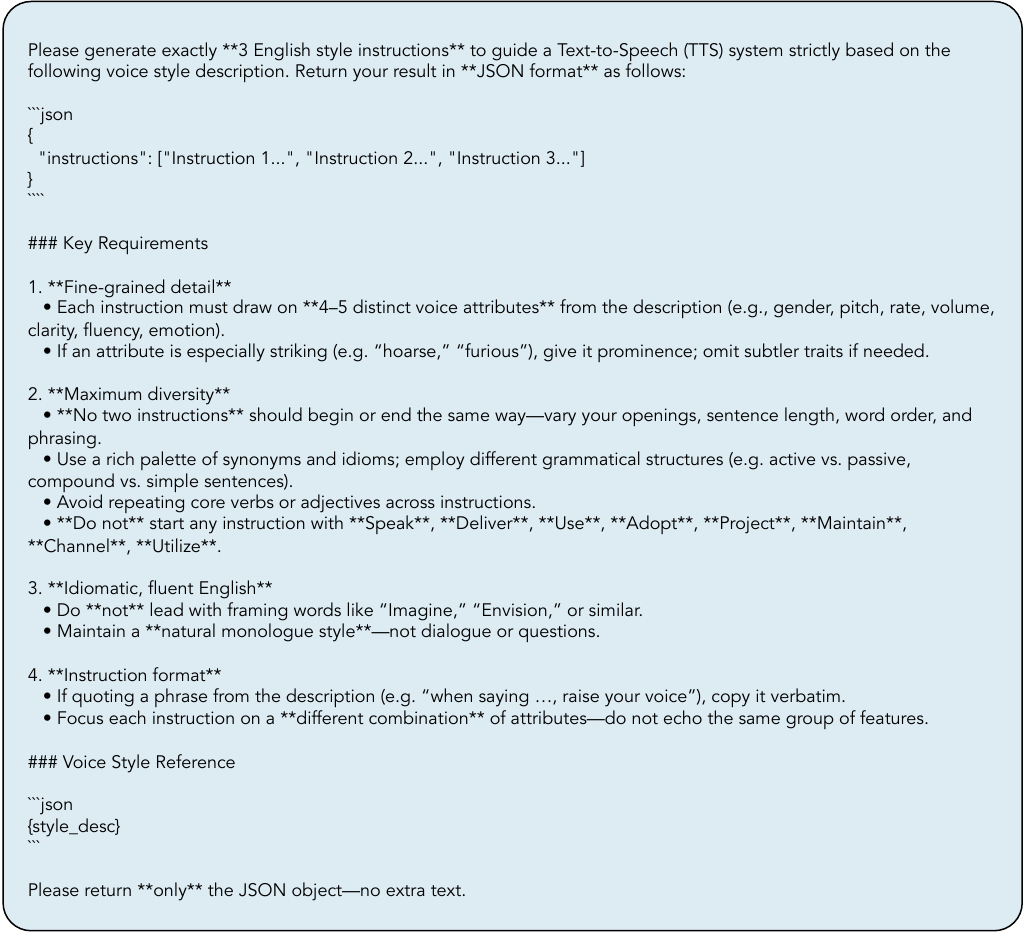}
    \caption{Prompts for RP instruction generation.}
    \label{fig:rp1}
\end{figure*}

\begin{figure*}
    \centering
    \includegraphics[width=\linewidth]{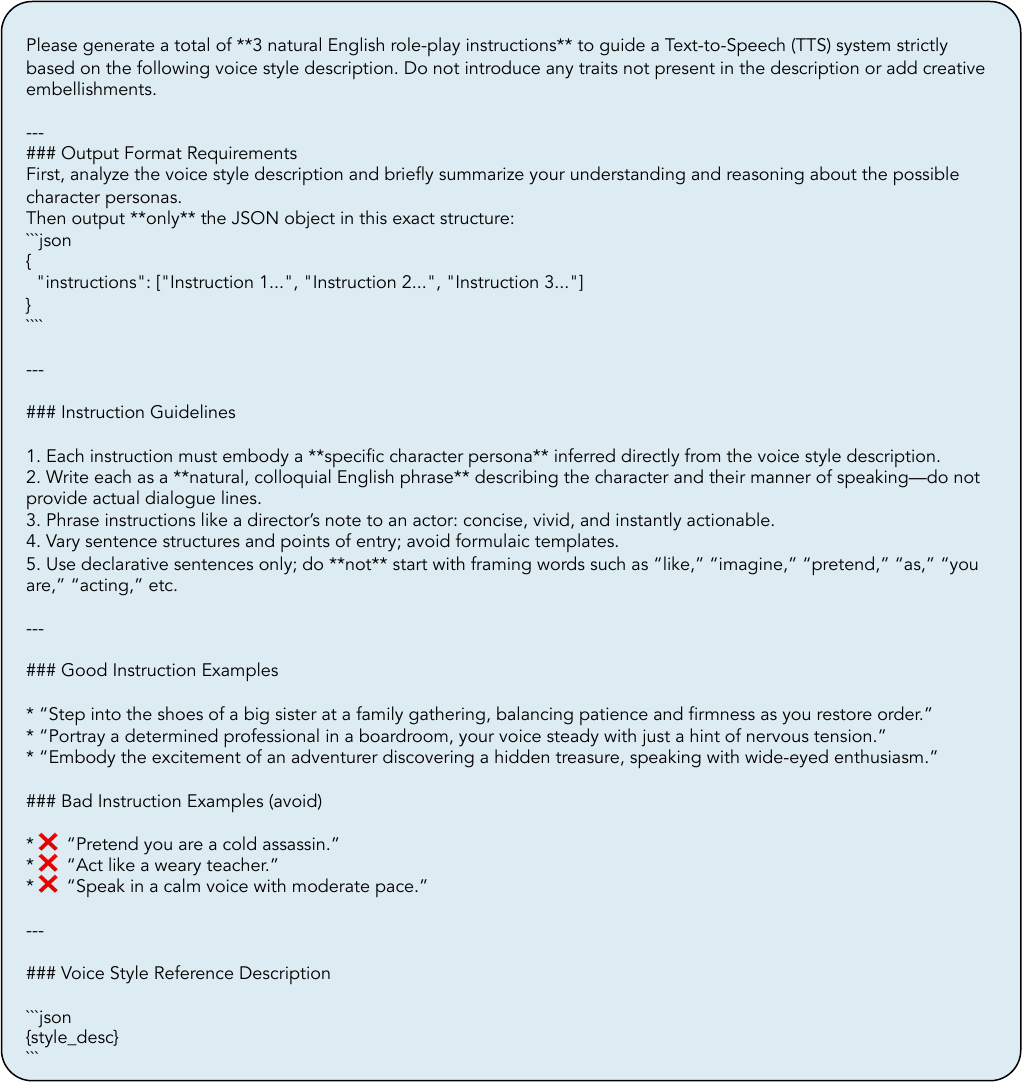}
    \caption{Prompts for RP instruction generation.}
    \label{fig:rp2}
\end{figure*}

\begin{figure*}
    \centering
    \includegraphics[width=\linewidth]{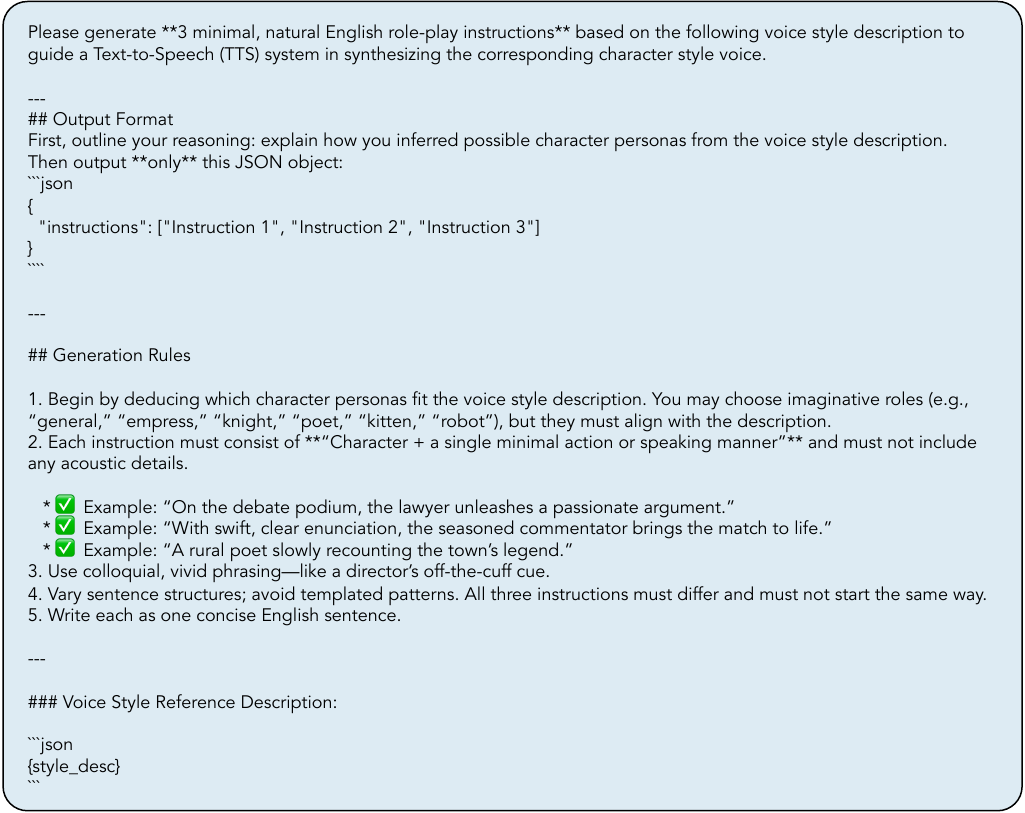}
    \caption{Prompts for RP instruction generation.}
    \label{fig:rp3}
\end{figure*}

To generate the DSD prompt, we design 3 weight settings to randomly drop out features, as shown in Fig.~\ref{tab:weights}. When instructing GPT-4o to generate instructions, we select a weight setting and provide the last features to the model. The prompt for generating DSD can be seen in Fig.~\ref{fig:dsd}. Also, we design 3 prompts for generating RP instructions, as shown in Fig.~\ref{fig:rp1} , Fig.~\ref{fig:rp2}, and Fig.~\ref{fig:rp3}.

\begin{table}[htbp]
    \centering
    \begin{tabular}{l  l l l}
        \toprule
                    & Set 1 & Set 2 & Set 3 \\
        \midrule
        probs       & 0.5     & 0.25   & 0.25  \\
        \midrule
        gender      & 1.0     & 0.5    & 0.5    \\
        pitch       & 1.0     & 0.5    & 0.5  \\
        speed       & 1.0     & 0.5    & 0.5  \\
        volume      & 1.0     & 0.5    & 0.5  \\
        age         & 1.0     & 0.8     & 0.5  \\
        clarity     & 1.0     & 0.5     & 0.5  \\
        fluency     & 1.0     & 0.5     & 0.5  \\
        accent      & 1.0     & 0.8     & 0.5  \\
        texture     & 1.0     & 0.8     & 0.5  \\
        emotion     & 1.0     & 1.0     & 0.8  \\
        tone        & 1.0     & 1.0     & 0.8  \\
        personality & 1.0     & 1.0     & 0.8  \\
        \bottomrule
    \end{tabular}
    \caption{Weight settings to randomly drop out features for DSD instructions.}
    \label{tab:weights}
\end{table}

\section{Scoring Guidelines}
\label{app:label}
\begin{figure*}
    \centering
    \includegraphics[width=\linewidth]{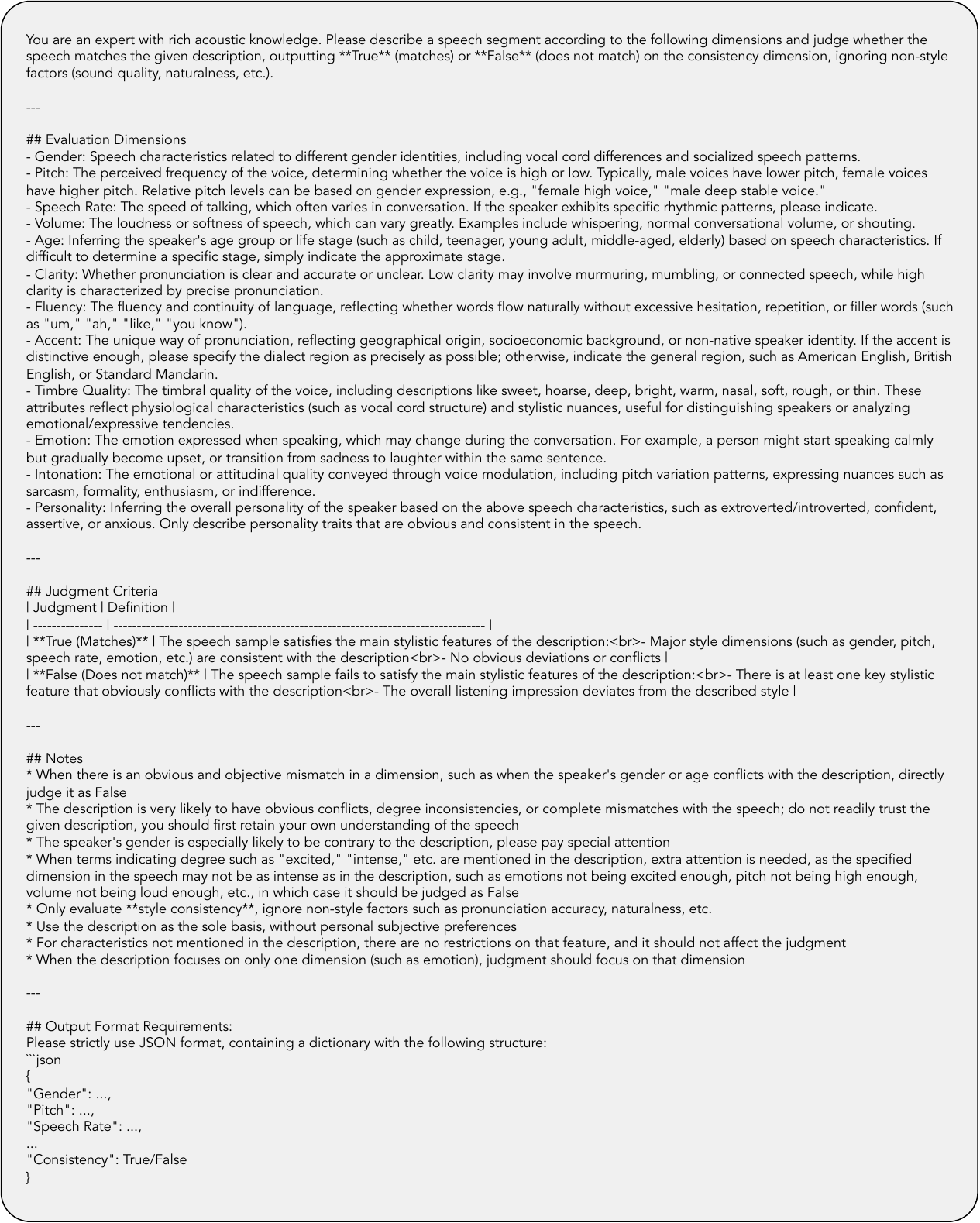}
    \caption{Prompt for Gemini-as-a-judge (translated ver.).}
    \label{fig:gemini_judge}
\end{figure*}

\begin{figure*}
    \centering
    \includegraphics[width=\linewidth]{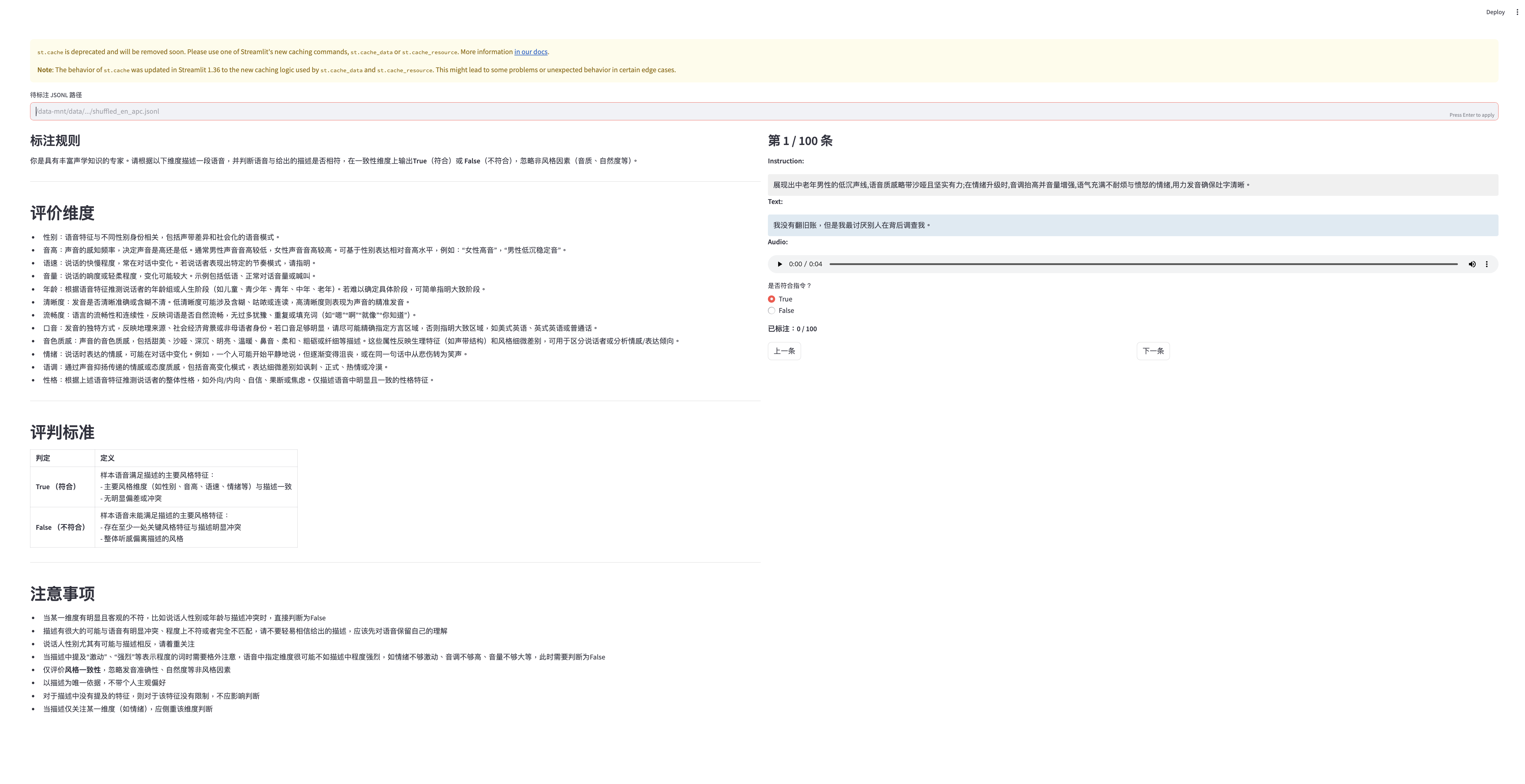}
    \caption{Screenshot for human annotation.}
    \label{fig:human_judge}
\end{figure*}

Fig.~\ref{fig:gemini_judge} illustrates our scoring guidelines for Gemini, and the screenshot of human annotation can be seen in Fig.~\ref{fig:human_judge}.

\end{document}